\documentclass[10pt]{article}

\usepackage{fullpage}
\usepackage{CJKutf8} 
\usepackage[utf8]{inputenc}
\usepackage{setspace}
\usepackage{parskip}
\usepackage{titlesec}
\usepackage[section]{placeins}
\usepackage{xcolor}
\usepackage{breakcites}
\usepackage{lineno}
\usepackage{hyphenat}
\usepackage{verbatim}

\PassOptionsToPackage{hyphens}{url}
\usepackage[colorlinks = true,
            linkcolor = blue,
            urlcolor  = blue,
            citecolor = blue,
            anchorcolor = blue]{hyperref}
\usepackage{etoolbox}

\usepackage{natbib}

\renewenvironment{abstract}
  {{\bfseries\noindent{\abstractname}\par\nobreak}\footnotesize}
  {\bigskip}

\titlespacing{\section}{0pt}{*3}{*1}
\titlespacing{\subsection}{0pt}{*2}{*0.5}
\titlespacing{\subsubsection}{0pt}{*1.5}{0pt}

\usepackage{authblk}

\usepackage{graphicx}
\usepackage[space]{grffile}
\usepackage{latexsym}
\usepackage{textcomp}
\usepackage{longtable}
\usepackage{tabulary}
\usepackage{booktabs,array,multirow}
\usepackage{amsfonts,amsmath,amssymb}
\providecommand\citet{\cite}
\providecommand\citep{\cite}

\newif\iflatexml\latexmlfalse

\AtBeginDocument{\DeclareGraphicsExtensions{.pdf,.PDF,.eps,.EPS,.png,.PNG,.tif,.TIF,.jpg,.JPG,.jpeg,.JPEG}}

\usepackage[utf8]{inputenc}
\usepackage[english,portuguese]{babel}

\usepackage{float}

\usepackage[margin=1.5in]{geometry}

\makeatletter
\providecommand{\bmsection}{\section} 
\renewcommand{\bibsection}{\section*{References}} 
\makeatother

\begin{document}

\title{NLP for Local Governance Meeting Records:\\ A Focus Article on Tasks, Datasets, Metrics and Benchmarks}

\author[1,2]{Ricardo Campos}

\author[2,3]{José Pedro Evans}

\author[2,3]{José Miguel Isidro}

\author[1,2]{Miguel Marques}

\author[2,3]{Luís Filipe Cunha}

\author[2,3]{Alípio Jorge}

\author[2,3]{Sérgio Nunes}

\author[2,3]{Nuno Guimarães}

\affil[1]{University of Beira Interior, Covilhã, Portugal}%
\affil[2]{INESC TEC, Porto, Portugal}%
\affil[3]{University of Porto, Porto, Portugal}%
\vspace{-1em}

  \date{}

\begingroup
\let\center\flushleft
\let\endcenter\endflushleft
\maketitle
\endgroup

\selectlanguage{english}
\begin{abstract}
Local governance meeting records are official documents, in the form of minutes or transcripts, documenting how proposals, discussions, and procedural actions unfold during institutional meetings. While generally structured, these documents are often dense, bureaucratic, and highly heterogeneous across municipalities, exhibiting significant variation in language, terminology, structure, and overall organization. This heterogeneity makes them difficult for non-experts to interpret and challenging for intelligent automated systems to process, limiting public transparency and civic engagement. To address these challenges, computational methods can be employed to structure and interpret such complex documents. In particular, Natural Language Processing (NLP) offers well-established methods that can enhance the accessibility and interpretability of governmental records. In this focus article, we review foundational NLP tasks that support the structuring of local governance meeting documents. Specifically, we review three core tasks: document segmentation, domain-specific entity extraction and automatic text summarization, which are essential for navigating lengthy deliberations, identifying political actors and personal information, and generating concise representations of complex decision-making processes. In reviewing these tasks, we discuss methodological approaches, evaluation metrics, and publicly available resources, while highlighting domain-specific challenges such as data scarcity, privacy constraints, and source variability. By synthesizing existing work across these foundational tasks, this article provides a structured overview of how NLP can enhance the structuring and accessibility of local governance meeting records.%
\end{abstract}%

\sloppy

\section{Introduction}
Local governance meeting records form the foundation of local democracy, documenting decisions on budgets, urban planning, and community regulations. These official records, whether in the form of minutes or transcripts, serve as the primary historical record of local government, supporting public accountability, civic engagement, and policy monitoring. Despite their relevance, they remain difficult to access for large-scale analysis \citep{brown2022councils}. Unlike national legislative data, such as Congress.gov or from the European Parliament \citep{erjavec2023parlamint}, which is often well-structured and standardized, local governance records are frequently fragmented, heterogeneous, and lack a standardized format \citep{brown2022councils}. They combine informal multi-speaker dialogue with formal administrative language, leading to dense, heterogeneous documents in which critical details, such as motions, decisions, and speaker positions, are often unintentionally obscured \citep{rodrigues2010knowledge}. This complexity hinders both human interpretation and computational analysis, limiting transparency and the potential for large-scale civic engagement. Current pre-trained models, however, struggle with domain-specific terminology of local government \citep{luz2018lener} thus motivating the development of specialized methods for effectively decomposing and analysis of governance documents \citep{kerkvliet-etal-2020-mentions, choi2023chatgpt}.


In this focus article, we review foundational NLP tasks that contribute to transforming raw local governance records into structured information. We organize the discussion around three core tasks that form the backbone of this process. Document Segmentation constitutes the first step, addressing the challenge of dividing long and heterogeneous meeting records into coherent, topic-based units ~\citep{hu2023meetingbank}, enabling targeted information retrieval, focused summarization, and the tracking of decisions across meetings. 
Building on this structure, domain-specific entity extraction enables the identification and classification of key entities such as political actors, institutions, locations, and domain-specific concepts embedded in governance discourse. Accordingly, we review Named Entity Recognition (NER) approaches not only in terms of standard entity categories, but also with respect to the fine-grained and domain-specific schemas required by municipal and administrative contexts. Finally, Automatic Text Summarization (ATS), addresses the challenge of condensing long meeting records into concise, accessible summaries, to allow readers to quickly understand key decisions and actions without the need to read full documents. 
Together, these three tasks illustrate how NLP techniques can be leveraged to move from unstructured local governance records toward structured representations that support analysis, retrieval, and enhance public access.

In the following sections, we discuss each task in detail, outlining methodological approaches, evaluation strategies, and domain-specific challenges, illustrating their relevance to the processing and interpretation of local governance meeting records.

\section{Document Segmentation}\label{Text_segmentation}

Local governance meeting records, including council minutes and transcripts, are typically long, weakly structured documents encompassing multiple agenda items, motions, and discussions within a single record. Their heterogeneity, combining formal administrative language with multi-speaker dialogue, makes it challenging to identify coherent topics and boundaries between discussions. Accurate segmentation of these records is essential for enabling downstream tasks in local governance, such as information retrieval, summarization and decision tracking. In the literature, this challenge is commonly addressed through document or text segmentation, which aims to partition a record into contiguous, semantically coherent units. To formally define this process, a document $D = \{u_1, u_2, \dots, u_n\}$ consisting of a sequence of atomic units  $u_i$ (e.g., sentences or utterances) is partitioned into a sequence of $k$ internally coherent segments $S = \{S_1, S_2, \dots, S_k\}$ \citep{hearst1997texttiling,galley2003discourse}.
Each unit $u_i$, can be labeled to indicate whether it represents the end of a segment, framing segmentation as a boundary detection problem \citep{koshorek2018textseg}.

Segmentation approaches can be broadly categorized into linear methods, producing a flat sequence of consecutive, non-overlapping segments, each representing a topically coherent unit \citep{ghinassi2024recent}, and hierarchical methods, which capture nested, multi-level structures \citep{eisenstein2009hierarchical}. In the context of local governance, hierarchical segmentation is particularly relevant, as a “Budget Approvals” section may contain sub-sections such as “Departmental Allocations”, “Amendment Proposals”, and “Final Vote Discussion”, reflecting the document’s topical structure.

Early research in text segmentation relied heavily on lexical-cohesion and surface-level features, drawing on classical cohesion theory \citep{halliday1976cohesion} and lexical chain models \citep{morrishirst1991lexical}. Methods such as TextTiling \citep{hearst1997texttiling}, C99 \citep{choi2000advances}, and graph-based approaches leveraging lexical similarity \citep{glavas2016unsupervised} exemplify this tradition. Lexical cues tend to perform well in contexts where topic continuity is expressed through repeated terminology (e.g., "budget allocation"), but their effectiveness degrades in the presence of stylistic variation, or heterogeneous discourse patterns, as is the case of local governance texts. Subsequent research expanded beyond surface-level lexical cues, introducing unsupervised strategies that leverage both local and global cohesion patterns. Window-based approaches, for instance, detect drops in lexical-similarity across adjacent blocks of text, revealing topical shifts \citep{hearst1997texttiling, galley2003discourse}. In contrast, Bayesian generative models, represent each segment as a latent multinomial distribution over words, inferring boundaries by maximizing the posterior probability of segment assignments \citep{eisenstein2008bayesian}. Unlike window-based methods, these probabilistic models perform global optimization, enabling coherent segmentation even when local lexical signals are weak, making them more robust to subtle shifts in topic common in council discussions. Clustering and graph-based approaches exploit sentence-level similarity or semantic graphs to group coherent regions \citep{choi2000advances,malioutov2006mincut,glavas2016unsupervised,riedl2012topictiling}, effectively capturing long-range dependencies, such as recurring discussions of the same motion throughout a meeting.

The emergence of deep learning shifted segmentation toward supervised boundary-prediction models, framing the task as a sequence labeling problem. Early neural approaches leveraged Bi-LSTMs to encode contextual dependencies across sentences, capturing dependencies that were not in the reach of traditional lexical or probabilistic models  \citep{koshorek2018textseg}, thus enabling better handling of multi-speaker dialogue and topic drift. Hierarchical neural models further integrated sentence and document-level features \citep{arnold2019sector}, allowing representations that reflect both local coherence and global document structure. More recently, transformer-based architectures including two-tier transformers \citep{glavas2020two} and BERT-based models \citep{lukasik2020textseg}, exploit self-attention mechanisms to model long-range discourse dependencies, achieving state-of-the-art segmentation performance. Recent approaches leveraging Large Language Models (LLMs) with Retrieval-Augmented Generation (RAG) \citep{nguyen2025enhancing} also show particular promise for local governance applications. By segmenting documents into semantically meaningful units and clustering them hierarchically, these methods support the retrieval of both low-level segments (e.g., a specific motion) and high-level clusters (e.g., an overarching policy debate), aligning naturally with the hierarchical structure of municipal records and facilitating more precise access for analysts, journalists, and citizens.

\section{Domain-specific Entity Extraction}
Identifying and classifying entities within local governance records is critical for a range of applications, including document anonymization, political accountability tracking, and information retrieval. These records frequently refer to persons (e.g., council members, citizens), organizations (e.g., municipal departments), locations (e.g., districts), and domain-specific entities (e.g., agenda items, votes, voters) reflecting the procedural and structured context of municipal meetings. Accurate recognition of such entities is therefore crucial for organizing records and enabling downstream NLP tasks. Named Entity Recognition (NER) provides, as a core NLP task, aimed at detecting and classifying entities into predefined categories, a formal framework for this process: given a token sequence $T = (t_1, t_2, \dots, t_N)$, the task consists of producing a set of tuples $(i_s, i_e, \ell)$, where $i_s$ and $i_e$ denote the start and end indices of an entity mention, and $\ell$ corresponds to a category label from a predefined set \citep{9039685}.
Despite its relevance, local government records have received little attention in NER research \citep{ramdhani2024named}, with only a few studies focusing on municipal or local contexts \citep{rodrigues2010knowledge, Peña2018OpenData, kimura-etal-2022-budget,citilink-dataset}. In contrast, substantial works exist on related domains such as parliamentary proceedings \citep{erjavec2023parlamint}, legal documents \citep{survey2025legal}, and administrative texts \citep{Reksi_NER}, which provide methodological and practical insights transferable to local contexts. This section reviews findings across these areas, focusing on span-level entity recognition, while leaving related tasks, such as coreference resolution, cross-document entity linking, argument mining, and stance detection, outside the scope of this article.

Entity categories in domain-specific NER vary according to the application \citep{marrero2013named}. In local government, the granularity of entity schemas depends on task requirements, ranging from coarse categories suitable for large-scale annotation \citep{erjavec2023parlamint} to fine-grained distinctions required for privacy protection\citep{garat2022automatic} or political accountability analysis \citep{kerkvliet-etal-2020-mentions}. For example, entity-based search and retrieval over local council documents benefit from standard categories such as Person, Location, Organization, and Time \citep{grishman-sundheim-1996-message,erjavec2023parlamint} enabling the identification of all references to a specific organization or track mentions of individuals across multiple minutes. Such coarse-grained schemes have been successfully applied in Finnish parliamentary speeches \citep{Tamper2022-FinParl} and Spanish open government data \citep{Peña2018OpenData}. 
  
Document Anonymization is another crucial application. Compliance with GDPR (General Data Protection Regulation)
(\citeyear{gdpr2016}) requires recognizing and masking sensitive information, which often extends beyond standard categories \citep{leitner2019fine, glaser2021anonymization, pilan2022text}. In European \citep{pilan2022text} and country-specific legal settings, NER is used to mask names, addresses, and health conditions, while preserving readability \citep{leitner2019fine,correia2022finegrained}.  Similarly, local governance records contain sensitive entities, such as citizen names, addresses, or personal circumstances appearing in licensing requests, or subsidy deliberations, demanding fine-grained masking \citep{leitner2019fine}.

Tracking political accountability constitutes a distinct application, requiring entity schemes that distinguish between individual mentions and role-based references. Various parliamentary corpora, including Portuguese \citep{albuquerque2022ulyssesner} and Dutch \citep{kerkvliet-etal-2020-mentions}, provide detailed entity annotations for political actors. The Public Entity Recognition (PER) framework \citep{nenno-NER-2025} further categorizes actors into Politicians, Parties, Authorities, Media, and Journalists, supporting cross-lingual political discourse analysis. 

Domain-specific entities are particularly relevant in local government records. Monetary expressions, subsidy amounts, and administrative actions are frequently absent in standard NER taxonomies but are critical in municipal records. Works in Portuguese municipal minutes \citep{rodrigues2010knowledge}, Japanese local assemblies \citep{kimura-etal-2022-budget}, focus on extracting these entities, linking financial items to deliberative context. Similarly, specialized categories, such as Metrics, Economics, and Binary Actions ~\citep{RuREBus-2021}, facilitate the analysis of regulatory and policy texts that are central to local governance.

Methodologically, NER in government contexts has evolved from hand-crafted rules \citep{mamede2016automated} and gazetteers \citep{cardoso-2012-rembrandt,rodrigues2010knowledge,Tamper2022-FinParl} to statistical sequence labeling models such as Conditional Random Fields (CRF) \citep{albuquerque2022ulyssesner,leitner2019fine,correia2022finegrained,Peña2018OpenData}, which better capture structured dependencies in government texts. Neural architectures subsequently improved recognition of complex contexts, ranging from efficient CNN/LSTM-based pipelines to large-scale Transformer models. Transformer-based approaches, including BERT variants (e.g., BERTimbau \citep{nunes-2024-named-ulysses} and FinBERT \citep{Tamper2022-FinParl}) now dominate the field due to their ability to capture rich contextual information, and are particularly effective for privacy-preserving masking, as seen in European court anonymization efforts \citep{pilan2022text}. Cross-lingual models such as XLM-R \citep{xlm-roberta-2020} have further enabled multilingual analysis of political corpora \citep{nenno-NER-2025}, while large-scale initiatives like ParlaMint \citep{erjavec2023parlamint} combine neural pipelines (Stanza \citep{qi-etal-2020-stanza}, Trankit \citep{nguyen-etal-2021-trankit}) for high-quality annotations. Finally, alternative contextualized representations, such as  fine-tuned ELMo models \citep{elmo-2018}, have been explored for administrative entity types \citep{RuREBus-2021}.

\section{Automatic Text Summarization}
Local governance meeting records are typically long, verbose, and procedurally dense, often spanning dozens or hundreds of pages. These documents interleave administrative procedures, formal motions, extended deliberations, and public interventions, making it difficult for non-expert readers to identify key decisions, actions, and outcomes, without reading full transcripts or minutes. As a result, even when records are publicly available, their length and complexity pose significant barriers to transparency, media scrutiny, and civic engagement. Automatic Text Summarization (ATS) addresses this challenge by producing concise representations ~\citep{RinoPardo2003} of meeting records that preserve their most salient information. In this context, summarization is not merely a generic text compression task, but a means of structuring institutional records around decisions and voting outcomes. Summaries are expected to surface agenda progression, highlight approved or rejected motions, and capture the rationale behind deliberations, thereby supporting public accountability and information access. 


Formally, text summarization can be defined as the task of generating a shorter text $S$ from an input document $D$, such that $S$ captures the most important content of $D$, while remaining coherent and readable. Approaches to summarization are generally classified as extractive ~\citep{Giarelis2023}, where salient sentences or segments are selected from the original document, or abstractive ~\citep{Giarelis2023}, where new sentences are generated to paraphrase and condense the source content. Hybrid approaches combine both extracted and generated text \citep{Aretoulaki1996, 10421343}. 

Beyond document-level summarization, local governance records are closely related to the broader field of meeting summarization \citep{Re-FRAME}, which has long served as a testbed for modeling complex, multi-speaker interactions. Meeting minutes themselves can be viewed as a specialized summary genre within the meeting domain, emphasizing the identification and structuring of key decisions. As such, advances in meeting summarization provide an important foundation for summarization in municipal and local governance settings.
Similar to other foundational NLP tasks, summarization techniques have evolved substantially over time from rule-based systems to data-driven neural architectures capable of learning complex language patterns ~\citep{gambhir2017survey, singh2021empirical, ZHANG2025131928, Liu2025-hallucination}. Rule-based approaches, often relying on manually engineered heuristics, struggled to generalize across document styles and institutional contexts  ~\citep{5392672, 10.1145/321510.321519, gambhir2017survey}.  This motivated the development of extractive approaches based on sentence scoring ~\citep{10.1145/1148170.1148269}, graph-based ranking algorithms such as Text Rank~\citep{mihalcea-tarau-2004-textrank}, and classification-based extractors that identify salient content units~\citep{10.1145/215206.215333}. While effective in constrained settings, these methods often fail to capture the procedural and discourse structure characteristic of governance records.

Abstractive summarization initially relied on sequence-to-sequence models ~\citep{rush-etal-2015-neural}, attention mechanisms ~\citep{nallapati-etal-2016-abstractive}, and pointer-generator networks ~\citep{gettothepoint}, which improved content selection and reduced redundancy. These developments culminated in Transformer-based models such as BART ~\citep{lewis2020bart} and T5 ~\citep{raffel2020exploring}, which substantially improved fluency and coherence ~\citep{singh2021empirical}. More recently, LLMs including GPT \citep{10.5555/3495724.3495883, 10.5555/3600270.3602281}, Gemini \citep{gemini2023}, and multilingual transformer models ~\citep{xiong-etal-2024-effective} have extended summarization capabilities through stronger reasoning, domain adaptation, and longer context windows ~\citep{ZHANG2025131928}. Hybrid models often incorporate reinforcement learning, coverage penalties, or constrained decoding strategies to improve factual consistency and content selection ~\citep{hokamp-liu-2017-lexically}.

Meeting-related domains introduce, however, additional challenges that go beyond those found in typical documents. These include multi-speaker dialogue, agenda-driven progress, frequent topic shifts, and domain-specific discourse phenomena such as turn-taking, action-items and long-range dependencies across interventations ~\citep{shriberg-etal-2004-icsi, carletta2005ami, qi-etal-2021-improving-abstractive, Liu2025DynamicAgenda}. Many of these characteristics are also present in local governance meetings, where deliberation, procedure, and public participation are integrated within the same record.

To address these challenges, specialized modeling strategies have been proposed for meeting summarization. These include dialogue segmentation techniques for long conversational documents ~\citep{chen-etal-2021-dialogsum}, speaker-aware encoders that explicitly model participant roles and contributions ~\citep{li-etal-2019-keep, zhong2021qmsum}, and approaches incorporating turn-taking patterns and temporal dependencies \citep{koay-etal-2021-sliding}. Such techniques are particularly relevant for municipal records, where accurately capturing who said what, when, and in relation to which agenda item is essential for producing informative and reliable summaries.

Domain-specific fine-tuning has consistently emerged as one of the most effective strategies, for adapting summarization  models to conversational and institutional contexts ~\citep{zhang2020pegasus, zhu-etal-2020-hierarchical, Jin2025CFAS}. Empirical studies show that models adapted to meeting or deliberative data yield summaries that are more coherent, informative, and decision-focused, with demonstrated success in settings such as council meetings, parliamentary proceedings, and collaborative discussions ~\citep{zhong2021qmsum}. These findings highlight the importance of domain adaptation for summarization in local governance, where generic summarization models often fail to capture procedural relevance and institutional nuance.

\section{Datasets}
The availability of annotated datasets specifically targeting local governance remains limited. While recent resources reflect a growing interest in the computational analysis of municipal-level data \citep{citilink-demo}, they predominantly focus on audiovisual records and spoken transcripts of local meetings, often without task-aligned linguistic annotations. Within this line of work, LocalView \citep{barari2023localview} provides a large collection of video recordings and transcripts of U.S. local government meetings, enabling research on political behavior and public participation, but without structured linguistic annotation. Similarly, van Wijk and Marx corpus \citep{van2025spoken} introduce a small dataset of transcribed Dutch municipal meetings, which captures the characteristics of spoken local governance but remains unannotated for downstream NLP tasks. More targeted supervision is offered by San Francisco City Council Media Coverage Dataset \citep{spangher2023tracking}, which annotates city council meeting transcripts with labels linking agenda items to media coverage.

Across the three tasks reviewed in this article, the availability of annotated datasets specifically targeting local governance remains limited. As a result, much of the methodological progress relies on resources developed for adjacent domains, such as Wikipedia, parliamentary proceedings, legal documents, and meeting corpora, which provide transferable supervision signals, annotation schemes, and evaluation benchmarks. Below, we summarize the most relevant resources for Document Segmentation, Domain-specific Entity Extraction, and Automatic Text Summarization, highlighting their relevance and limitations for local governance applications.

In Document Segmentation, publicly available datasets derived from local governance records are scarce. Consequently, research in this area has largely relied on large-scale, open-domain benchmarks. Two large-scale resources have become central for training and evaluating segmentation models. WikiSection \citep{koshorek2018textseg} provides crowd-verified segment boundaries for approximately 40k English and German Wikipedia articles spanning domains such as cities, celebrities, diseases, and organisms. Each article is paired with its human-written section structure, yielding high-quality hierarchical topic boundaries that support both coarse-grained and fine-grained segmentation evaluation. Complementing this, Wiki-727K \citep{koshorek2018textseg} offers a substantially larger collection of 727K English Wikipedia articles, where section boundaries are automatically aligned with headings to create weakly supervised segmentation labels. While noisier, Wiki-727K enables training segmentation models at scale and exposes them to stylistic and structural variability.Although these datasets do not reflect the nature of local governance meetings, they form the backbone of current neural segmentation research, providing section-aware annotations at multiple levels of granularity that are difficult to obtain for municipal records. As such, they are commonly used for pretraining or methodological benchmarking before adaptation to governance-related data.


A similar scarcity characterizes NER resources for local governance, directing research toward parliamentary, legal, and administrative corpora that share institutional language and accountability requirements. In the parliamentary domain, ParlaMint \citep{erjavec2023parlamint} constitutes a foundational multilingual benchmark, encompassing approximately 500 million words across 17 European parliaments, annotated with standard entity categories. More fine-grained modeling of political actors is supported by the Dutch Parliamentary corpus \citep{kerkvliet-etal-2020-mentions} which introduces a 12-type ontology capturing both named entities and role-based descriptions, such as ministers and committees. Similarly, UlyssesNER-BR \citep{albuquerque2022ulyssesner} annotates Brazilian legislative documents with fine-grained person subtypes, disentangling individuals from their institutional roles, an issue directly relevant to municipal decision-making records. In the legal domain, datasets such as the German Legal Documents corpus \citep{leitner2019fine} and the Brazilian Supreme Court corpus \citep{correia2022finegrained} employ hierarchical annotation schemes with 19 and 24 nested entity types, respectively, enabling precise extraction of sensitive information and institutional actors. Privacy-oriented benchmarks, including the Text Anonymization Benchmark \citep{pilan2022text} focus on detecting quasi-identifiers for privacy preservation. Finally, RuREBus \citep{RuREBus-2021} addresses administrative governance directly, introducing specialized categories, such as Metrics and Economics, that are absent from standard NER benchmarks but crucial for regulatory and policy analysis in local government contexts.


For Automatic Text Summarization, several datasets provide partial or indirect coverage of local governance phenomena, particularly through meeting summarization. MeetingBank \citep{hu2023meetingbank}, aligns long city council transcripts with human-written summaries, offering one of the closest domain matches for municipal summarization, although it lacks explicit ground truth for segmentation. Other resources emphasize query-focused or topic-structured summarization. For example, QMSum \citep{zhong2021qmsum} provides hierarchical annotations linked to decisions, actions and agenda items, enabling fine-grained retrieval and summarization within long meetings. Datasets of legislative and assembly minutes further support abstractive summarization at the topic level. The Japanese Regional Assembly Minutes Corpus \citep{Shirafuji2020, kimura-etal-2016-creating, kimura2020ntcir15} and the Czech and English parliamentary dataset \citep{nedoluzhko-etal-2022-elitr} provide structured summaries enriched with metadata, reflecting formal deliberative settings similar to local councils. Finally, dialogue-centric meeting corpora, such as ICSI \citep{shriberg-etal-2004-icsi} and AMI \citep{AMI_meeting_corpus, carletta2006ami} have played a central role in advancing techniques for hierarchical, multi-party summarization, as they include detailed speaker and discourse annotations. 
Complementing these task-specific and adjacent-domain resources, the recently proposed CitiLink-Minutes \citep{citilink-dataset} introduces a distinct type of local governance dataset centered on official written meeting minutes. The dataset comprises 120 municipal meeting minutes from six Portuguese municipalities and provides multilayer annotations spanning personal information, metadata, subjects of discussion, summarization, voting, and outcomes. Unlike prior datasets that typically address individual NLP tasks in isolation, CitiLink-Minutes adopts a task-agnostic, document-level annotation strategy supporting a broad range of downstream applications, including—but not limited to segmentation, entity extraction, and summarization. 
Table \ref{tab:resources} summarizes the key characteristics of the datasets discussed across the three tasks, highlighting their domains, scale, and relevance for local governance research.

\begin{table}[t]
\caption{In-domain and adjacent-domain datasets used for segmentation, entity extraction, and summarization of meeting-centric records.}
\label{tab:resources}
\centering
\resizebox{\textwidth}{!}{%
\begin{tabular}{|l|l|c|l|l|l|l|l|}
\hline
\multicolumn{1}{|c|}{\textbf{Dataset}} &
\multicolumn{1}{c|}{\textbf{Domain}} &
\multicolumn{1}{c|}{\textbf{Language}} &
\multicolumn{1}{c|}{\textbf{Docs}} &
\multicolumn{1}{c|}{\textbf{Annotations}} &
\multicolumn{1}{c|}{\textbf{Avg. Length (Tok)}} &
\multicolumn{1}{c|}{\textbf{Task(s)}} &
\multicolumn{1}{c|}{\textbf{Observations}} \\ \hline

\textbf{WikiSection} &
Wikipedia &
EN, DE &
38,000 &
242,000 &
$\sim$3,500 &
Text Segmentation &
\begin{tabular}[c]{@{}l@{}}Well-structured articles\\Strong baseline for linear segmentation.\end{tabular} \\ \hline

\textbf{Wiki-727K} &
Wikipedia &
EN &
727,746 &
$\sim$3.5M &
$\sim$2,000 &
Text Segmentation &
\begin{tabular}[c]{@{}l@{}}Massive scale enables training of hierarchical models\\Clean structure unlike real meetings.\end{tabular} \\ \hline

\textbf{ParlaMint} &
Parliamentary &
MULTI (17) &
17 corpora &
-- &
-- &
NER &
4 standard entity types (PER, LOC, ORG, MISC). \\ \hline

\textbf{UlyssesNER-BR} &
Legislative &
PT-BR &
950 &
9,526 &
138,741 &
NER &
Fine-grained person subtypes (individual vs. role). \\ \hline

\textbf{German Legal Documents} &
Legal (Courts) &
DE &
750 &
66,723 &
2,157,048 &
NER &
19 fine-grained types subdividing Person and Location. \\ \hline

\textbf{Brazilian Supreme Court} &
Legal (Courts) &
PT-BR &
594 &
62,933 &
1,780,000 &
NER &
4 coarse + 24 nested fine-grained types. \\ \hline

\textbf{Text Anonymization Benchmark} &
Legal (ECHR) &
EN &
1,268 &
155,006 &
1,828,970 &
NER &
Privacy-oriented evaluation. \\ \hline

\textbf{RuREBus} &
Administrative &
RU &
-- &
120,989 &
394,966 &
NER &
8 specialized entity types (Metrics, Economics, etc.). \\ \hline

\textbf{Dutch Parliamentary} &
Parliamentary &
NL &
5,536 &
3,579 &
86,206 &
NER &
\begin{tabular}[c]{@{}l@{}}12 entity types with complex descriptions\\(e.g., ``the Minister of Education'').\end{tabular} \\ \hline

\textbf{MeetingBank} &
City Council &
EN &
1,366 &
6,892 &
$\sim$28,000 &
Summarization &
\begin{tabular}[c]{@{}l@{}}Real long-form council meetings\\Closest domain match but lacks ground truth for segmentation.\end{tabular} \\ \hline

\textbf{QMSum} &
Mixed Meetings &
EN &
232 &
1,808 &
$\sim$10,000 &
Summarization &
Fine-grained query-relevant spans rather than contiguous segments. \\ \hline

\textbf{Tokyo Assembly Minutes} &
Assembly Minutes &
JA &
-- &
$\sim$27,000 &
-- &
Summarization &
\begin{tabular}[c]{@{}l@{}}Formal document-style minutes\\Structured deliberation records.\end{tabular} \\ \hline

\textbf{ELITR (English)} &
\begin{tabular}[c]{@{}l@{}}Parliamentary\\Project Meetings\end{tabular} &
EN &
120 &
-- &
$\sim$7,000 &
Summarization &
\begin{tabular}[c]{@{}l@{}}Multi-level summaries\\Multi-level corpora.\end{tabular} \\ \hline

\textbf{ELITR (Czech)} &
\begin{tabular}[c]{@{}l@{}}Parliamentary\\Project Meetings\end{tabular} &
CS &
59 &
-- &
$\sim$8,500 &
Summarization &
\begin{tabular}[c]{@{}l@{}}Multi-level summaries\\Multi-level corpora.\end{tabular} \\ \hline

\textbf{ICSI} &
Academic Meetings &
EN &
75 &
-- &
$\sim$11,000 &
Summarization &
Includes speaker turns, dialogue structure, and annotations. \\ \hline

\textbf{AMI} &
Design Meetings &
EN &
137 &
-- &
$\sim$7,000 &
Summarization &
Rich metadata: roles, dialogue acts, multimodal recordings. \\ \hline

\textbf{CitiLink-Minutes} &
Municipal Meeting Minutes &
PT-PT &
120 &
-- &
$\sim$1,000,000 &
Multitask &
\begin{tabular}[c]{@{}l@{}}Recently proposed multilayer dataset.\\Covers multiple layers such as PI, metadata,\\ subjects, summarization, voting and outcomes.\end{tabular} \\ \hline

\end{tabular}%
}
\end{table}

\section{Metrics and Benchmarks}


Evaluating the success of extracting relevant information from unstructured text in the domain of local governance meeting records requires a diverse set of metrics that go beyond generic text benchmarks. In this section, we review the evaluation strategies commonly adopted for the three NLP tasks discussed in this focus article.

Text segmentation evaluation focuses on assessing the accurate identification of topical boundaries within a document. Exact-match metrics are often too restrictive for long and heterogeneous records, where boundary placements that deviate slightly from the reference may still be functionally appropriate. To address this, segmentation quality is typically measured using probabilistic metrics that tolerate near-boundary errors, such as the $P_k$ score \citep{beeferman1997} and WindowDiff \citep{pevznerhearst2002critique}. The $P_k$ metric estimates the likelihood that two textual units sampled $k$ positions apart are incorrectly classified as belonging to the same or different segment, with lower scores indicating better performance. WindowDiff \citep{pevznerhearst2002critique} refines this approach by sliding a fixed-size window across the document and penalizing discrepancies between the number of predicted and ground truth boundaries within each window, making it more sensitive to over and under-segmentation. More recently, Boundary Similarity ($B$) \citep{fournier2013evaluating} has been proposed as an edit-distance-based measure that assigns partial credit to boundaries placed close to the gold standard, offering a more nuanced evaluation when compared to $P_k$ or WindowDiff. These evaluation measures are particularly well-suited to council minutes and meeting transcripts, where deliberations often drift, recur, or overlap across agenda items, making strict boundary alignment unrealistic for practical use.

Across the WikiSection benchmark \citep{koshorek2018textseg}, segmentation performance is predominantly reported using the $P_k$ metric, which has become the de facto standard for quantifying segmentation error. The current state-of-the-art, introduced by \citep{yu2023improvinglong}, reports a 4.3\% relative reduction in $P_k$ over the Longformer \citep{Beltagy2020Longformer} baseline, primarily by optimizing coherence modeling in long-context representations. Earlier foundational work, such as the Hierarchical BiLSTM architecture by Koshorek et al. \citep{koshorek2018textseg}, reported only segmentation-level Precision, Recall, and F1 scores. However, the field has since converged on probabilistic error metrics, including $P_k$ and WindowDiff, which better capture the practical utility of near-boundary predictions. This shift reflects growing recognition that strict boundary matching fails to reward segments that are functionally correct but not perfectly aligned with reference indices.

A similar evaluation practice is observed in the Wiki-727K corpus \citep{koshorek2018textseg}, where $P_k$ has served as the primary benchmark metric since its introduction. On this dataset, \citep{yu2023improvinglong} reports the current best performance with a $P_k$ of 13.89, improving upon the previous baseline of 15.0. While earlier supervised approaches, including the LSTM-based model of ~\citep{koshorek2018textseg} and the Cross-Segment Attention transformer proposed by ~\citep{lukasik2020textseg}, emphasized segmentation F1, modern comparisons prioritize $P_k$ to ensure standardized evaluation and comparability across segmentation studies.


For domain-specific entity extraction, evaluation traditionally relies on precision, recall, and F1 at the entity level, typically requiring exact boundary and type matches \citep{conll-2003-introduction}. While this strict criterion is appropriate for well-defined textual domains, it can be overly restrictive in local governance contexts. In municipal records, partial recognitions, such as correctly identifying a politician’s name while misclassifying their role or affiliation, may still support downstream applications, including search or indexing. To address this limitation, relaxed evaluation schemes have been proposed, assigning partial credit to boundary mismatches or type confusions within hierarchical entity taxonomies \citep{segura-bedmar-etal-2013-semeval}. Evaluation priorities also vary by application. In anonymization-oriented NER, recall is often weighted more heavily than precision, as failing to identify a single personal identifier can result in a privacy breach, whereas over-masking is typically more acceptable \citep{pilan2022text}.

Empirical results across governmental and administrative domains show a clear performance progression as modeling approaches have evolved. Early rule-based systems reported F1 scores ranging from 63\% to 89\% \citep{cardoso-2012-rembrandt,mamede2016automated}, followed by CRF-based approaches achieving between 76\% and 93\% \citep{albuquerque2022ulyssesner,leitner2019fine,luz2018lener}. More recently, transformer-based models have pushed performance further, reaching 82\% to 96\% F1 across several benchmarks \citep{leitner2019fine,nunes-2024-named-ulysses,pilan2022text}. However, performance varies substantially with the specificity and complexity of the entity schema: well-established categories used in legal anonymization consistently exceed 90\% F1 \citep{leitner2019fine,pilan2022text}, whereas fine-grained administrative entity schemes, remain considerably more challenging, with reported F1 scores between 57\% and 76\% \citep{RuREBus-2021}.


Finally, evaluation in automatic text summarization relies on specialized metrics designed to compare generated summaries with reference texts, traditionally grouped into lexical and semantic approaches \citep{10.1162/089120102762671927}. Lexical metrics measure surface-level overlap, typically through n-gram overlap or token matching ~\citep{fabbri-2024-summeval, peyrard-2019-studying}. Widely used measures such as ROUGE ~\citep{lin-2004-rouge}, Moverscore ~\citep{zhao-etal-2019-moverscore}, METEOR \citep{banerjee-lavie-2005-meteor} or BLEU ~\citep{papineni-etal-2002-bleu}, differ in how they capture coverage, alignment, and lexical variation. While effective for measuring content overlap, these metrics can be insensitive to assess whether summaries accurately reflect decisions, actions, and procedural outcomes that are central to local governance records. To address these limitations, semantic metrics leverage contextualized representations to estimate similarity at the meaning level. Approaches such as BERTScore ~\citep{zhao-etal-2023-discoscore} and other embedding-based distances ~\citep{pmlr-v37-kusnerb15, Reimers2019SentenceBERTSE} better account for paraphrasing which are common in summaries of deliberative meetings. Nevertheless, these metrics remain limited in evaluating whether summaries accurately capture decision rationales, voting outcomes, or agenda progression, making human evaluation essential for assessing fluency, coherence, redundancy, factuality, and informativeness in governance summaries ~\citep{mani2001summarization}.

Empirical results reflect the variability and difficulty of summarizing governance-related meetings. On the MeetingBank dataset, which aligns city council transcripts with human-written summaries, systems report ROUGE-1 scores between 62-70 and ROUGE-2 between 51-60, alongside strong embedding-based and QA-based evaluation with BERTScore ranging  77-82 and MoverScore from 66-71 ~\citep{hu2023meetingbank}. In contrast, results on the ELITR English dataset \citep{nedoluzhko-etal-2022-elitr} are substantially lower, with ROUGE-1 scores between 35-42, ROUGE-2 between 5-10, BERTScore between 46-52, and BLEU between 5-6 ~\citep{nedoluzhko-etal-2022-elitr}, highlighting the increased difficulty posed by heterogeneous meeting styles and weaker alignment between transcripts and summaries. Language variation further exacerbates these challenges, with a ROUGE-1 of 6.96 reported for Japanese assembly minutes \citep{Shirafuji2020}. On the QMSum \citep{zhong2021qmsum}, which targets decision and query-focused summaries, the best abstractive models achieve ROUGE-1 scores of 32-36, ROUGE-2 of 8-11, and ROUGE-L of 28-31. Overall, these results suggest that while meaningful progress has been made, they remain insufficient to fully address the complexity and informational needs of local governance meetings.  

In addition to the task-specific results reviewed above, the CitiLink-Minutes dataset \citep{citilink-dataset} provides baselines for structured information extraction, voting identification, and topic classification on municipal meeting minutes, using both encoder-based and generative few-shot models as reproducible reference points for future research.

\section{Conclusion and Future Work}

This focus article examined how three core NLP tasks, Document Segmentation, Domain-specific Entity Extraction, and Automatic Text Summarization, contribute to structuring and interpreting local governance meeting records. Across these tasks, we showed that although substantial methodological progress has been achieved in adjacent domains, local governance remains comparatively underexplored, despite its societal relevance and increasing availability of public data. A central conclusion of this review is that the main obstacles in this domain are not algorithmic but structural. The scarcity of annotated municipal corpora, and the limited suitability of existing evaluation practices collectively constrain progress. As a result, models trained or benchmarked on news, legal, or parliamentary data often fail to generalize to the procedural, multi-speaker, and decision-oriented nature of local governance records. Several open challenges emerge from this analysis. First, there is a pressing need for annotated, multilingual, publicly available datasets of meeting transcripts and official minutes. Second, task formulations and evaluation metrics must move beyond news-centric assumptions to better reflect the realities of municipal meetings. Finally, while LLMs offer promising avenues for low-resource settings, their adoption in local governance requires careful validation against domain-specific benchmarks that are largely absent today. Addressing these challenges is essential for advancing NLP methods that meaningfully support transparency, accountability, and civic engagement at the local level.

\section*{Funding Information}

{\label{974317}}

This work was financed within the scope of the project  CitiLink, with reference 2024.07509.IACDC, which is co-funded by Component 5 - Capitalization and Business Innovation, integrated in the Resilience Dimension of the Recovery and Resilience Plan within the scope of the Recovery and Resilience Mechanism (MRR) of the European Union (EU), framed in the Next Generation EU, for the period 2021 - 2026, measure RE-C05-i08.M04 - "To support the launch of a programme of R\&D projects geared towards the development and implementation of advanced cybersecurity, artificial intelligence and data science systems in public administration, as well as a scientific training programme," as part of the funding contract signed between the Recovering Portugal Mission Structure (EMRP) and the FCT - Fundação para a Ciência e a Tecnologia, I.P. (Portuguese Foundation for Science and Technology), as intermediary beneficiary.{\url{https://doi.org/10.54499/2024.07509.IACDC}}

\section*{Author Contributions}

\textbf{Ricardo Campos}: Conceptualization, Writing-Original Draft, Writing - Review \& Editing, Supervision, Project Administration, Funding Acquisition
\textbf{José Pedro Evans}: Investigation, Writing-Original Draft
\textbf{José Miguel Isidro}: Investigation, Writing-Original Draft
\textbf{Miguel Marques}: Investigation, Writing-Original Draft
\textbf{Luís Filipe Cunha}: Writing - Review \& Editing, Supervision
\textbf{Sérgio Nunes}: Writing - Review \& Editing, Supervision, Project Administration
\textbf{Alípio Jorge}: Writing - Review \& Editing, Supervision, Project Administration
\textbf{Nuno Guimarães}: Conceptualization, Methodology, Writing - Review \& Editing, Supervision, Project Administration

\section*{Generative AI Usage}
This article was written with the assistance of generative artificial intelligence to enhance clarity and readability. The AI tools were used solely to improve language flow and presentation.
\FloatBarrier


\bibliographystyle{apalike}

\bibliography{wileyNJD-AMA}

\end{document}